\documentclass{bmvc2k}

\title{Two Heads are Better than One: Geometric-Latent Attention for Point Cloud Classification and Segmentation}

\addauthor{Hanz Cuevas-Velasquez}{hanz.c.v@ed.ac.uk}{1}
\addauthor{Antonio Javier Gallego}{jgallego@dlsi.ua.es}{2}
\addauthor{Robert B. Fisher}{rbf@inf.ed.ac.uk}{1}

\addinstitution{
 School of Informatics \\
 University of Edinburgh \\
 Edinburgh, UK
}
\addinstitution{
 Department of Software and Computing Systems\\
 University of Alicante, \\
 Alicante, Spain
}

\runninghead{Cuevas-Velasquez, Fisher, Gallego}{Geometric-Latent Attention}

\def\etal{\emph{et al}\bmvaOneDot}

\usepackage{amssymb}
\usepackage{multirow}
\usepackage[inline]{enumitem}
\newcommand{\review}[1]{#1}
\begin{document}
\setlength{\belowdisplayskip}{2pt} \setlength{\belowdisplayshortskip}{2pt}
\setlength{\abovedisplayskip}{2pt} \setlength{\abovedisplayshortskip}{2pt}

\maketitle

\begin{abstract}
We present an innovative two-headed attention layer that combines geometric and latent features to segment a 3D scene into semantically meaningful subsets. Each head combines local and global information, using either the geometric or latent features, of a neighborhood of points and uses this information to learn better local relationships. This Geometric-Latent attention layer (Ge-Latto) is combined with a sub-sampling strategy to capture global features. Our method is invariant to permutation thanks to the use of shared-MLP layers, and it can also be used with point clouds with varying densities because the local attention layer does not depend on the neighbor order. Our proposal is simple yet robust, which allows it to achieve competitive results in the ShapeNetPart and ModelNet40 datasets, and the state-of-the-art when segmenting the complex dataset S3DIS, with 69.2\% IoU on Area 5, and 89.7\% \review{overall accuracy} using K-fold cross-validation on the 6 areas.

%The simplicity of the method allows it to achieve better results even when trained with a smaller number of points than the current state-of-the-art methods.

%This document demonstrates the format requirements for papers submitted
%to the British Machine Vision Conference.  The format is designed for
%easy on-screen reading, and to print well at one or two pages per sheet.
%Additional features include: pop-up annotations for
%citations~\cite{Authors06,Mermin89}; a margin ruler for reviewing; and a
%greatly simplified way of entering multiple authors and institutions.

%{\bf All authors are encouraged to read this document}, even if you have
%written many papers before.  As well as a description of the format, the
%document contains many instructions relating to formatting problems and
%errors that are common even in the work of authors who {\em have}
%written many papers before.
\end{abstract}

%-------------------------------------------------------------------------
\section{Introduction}
\label{sec:intro}

Robotics, autonomous driving, and related areas rely heavily on information captured by 3D sensors like RGB-D cameras, stereo cameras, and LiDARs. This information provides to the agent (robots or cars) the 3D location of their surroundings, which can be processed and used in tasks like scene understanding, path planning, navigation, among others \cite{cuevas2018hybrid, cuevas2020real, strisciuglio2018trimbot2020}. %(removed) One way to find the objects in the 3D space and their location is through point cloud segmentation. 
\review{One of the most used approaches to detect objects in 3D space is point cloud segmentation.} A point cloud is a set of points in 3D, usually unordered and sparse; some regions can be densely populated and others empty. This type of non-grid structured data is difficult to be used with convolution operators with the same efficiency as their 2D counterpart. 
%The 3D data captured by these devices comes in the form of point clouds.

Various approaches have been proposed to handle such data. Some approaches project the 3D raw data into a regular structure (e.g. voxels) where 3D convolutions can be used \cite{zhou2018voxelnet, meng2019vv, riegler2017octnet, wang2017cnn, klokov2017escape, shao2018h}. Other approaches use multilayer perceptrons (MLP) to process point clouds directly  \cite{qi2017pointnet, qi2017pointnet++, thomas2019kpconv}. %using as a backbone pointNet.
A third approach is to project the points to an intermediate grid structure where 2D convolutions can be used \cite{lin2020fpconv, huang2019texturenet, yang2020pfcnn}. Lately, with the success of transformers and attention mechanisms in the area of natural language processing (NLP) \cite{vaswani2017attention}, these methods %have started to show dominance in this area \cite{guo2020pct, qiu2021semantic}.
\review{are starting to be used for 3D point cloud problems \cite{guo2020pct, qiu2021semantic}.}

\textbf{This paper proposes a multi-head attention layer called Geometric-Latent Attention (Ge-Latto) to segment and label subsets of the point cloud}. Ge-Latto is a two-headed local attention layer that evaluates a patch inside the point cloud and tries to find good relationships between the neighbor points. Unlike other works that combine all the features indiscriminately \cite{zhao2020exploring, li2018pointcnn}, each attention head focuses on a specific type of feature. One head is in charge of finding good geometric relations and the other in finding relationships among the latent features of the network. 
Similar to \cite{lin2020fpconv}, we use an encoder-decoder network with residual connections. Each layer of the encoder sub-samples the input points, groups the points into neighborhoods, and uses our Ge-Latto layer to find local-spatial relationships from the latent and geometric features of neighbor points. The neighbors are found using radius neighborhoods instead of k-nearest-neighbors (kNN). The network increases this radius in each layer to increase the field of view and find relationships in bigger neighborhoods. In the decoder part, we up-sample the points using tri-linear interpolation. To ensure that in each sampled layer the network learns useful features, we add auxiliary losses similar to PSPNet \cite{zhao2017pyramid} and RetinaNet \cite{lin2017focal}. In other words, the network predicts the segmentation for each sample size as seen in Figure \ref{fig:network}. Our approach is also invariant to permutation because all the layers are shared MLPs. 

\textbf{The main contributions of this paper are:} %in bold because of the correction
\begin{enumerate*}[label=\textbf{\alph*)}]
    \item A novel two-headed attention layer that is able to combine efficiently the geometric and latent information of unordered point clouds with variable densities for semantic segmentation \review{and shape classification}. 
    \item A pyramid-based encoder-decoder architecture with \review{multi-resolution outputs and} auxiliary losses to leverage feature patterns at different resolutions.
    \item State of the art performance on the complex dataset S3DIS. Not only in the area 5, but also in the k-fold cross-validation overall accuracy, as well as competitive results in the ShapeNetPart and ModelNet40 datasets.
\end{enumerate*}
%In the encoder part, before passing the information from one layer to another, the points are sub-sampled. To find local-spatial relationships, the points are grouped in neighborhoods and processed by our Ge-Latto layer. To obtain the neighbors of a point, the network uses radius neighborhoods instead of k-nearest-neighbors (kNN). The network is able to find relationships in bigger neighborhoods by increasing the field of view every time a new sample is taken. This is done by increasing the grouping radius at every encoder layer. In the decoder part, we up-sample the points using tri-linear interpolation. To ensure that in each sampled layer the network learns useful features, we add auxiliary losses. In other words, the network predicts the segmentation for each sample size as seen in Figure \ref{fig:network}. Our approach is also invariant to permutation because all the layers are shared MLPs.

\section{Related Work}
\label{sec:rel_work}
Recent works focus on how to handle unordered 3D points and find spatial relationships between them to better segment a point cloud. This section briefly reviews these methods and groups them into 4 categories. %This section briefly review previous methods to segment point clouds and groups them into 3 categories.

\noindent \textbf{Volumetric-based methods.}
These methods quantize an unordered point cloud in a uniform structure like voxels. Some approaches use 3D convolutions to find local relationships between closer groups of points \cite{zhou2018voxelnet, meng2019vv}. However, the amount of memory required to compute these convolutions makes them unfeasible to process a large number of points. Methods like OctNet \cite{riegler2017octnet} and O-CNN \cite{wang2017cnn} save computation time by using octrees to avoid processing empty spaces. \cite{klokov2017escape} and \cite{shao2018h} use Kd-tree and Hash structures instead. \review{\cite{tang2020searching} uses sparse 3D convolutions rather than efficient data structures.} Although these implementations reduce the computation required to train a 3D CNN, quantizing the points comes with the cost of losing important fine-grained information.

\noindent \textbf{Point-based networks.}
These are networks capable of using irregular point clouds without projecting or quantizing them into regular grids. Their main characteristic is the use of shared MLP layers, also known as point-wise or 1D convolutional layers. PointNet \cite{qi2017pointnet} is a milestone of this kind of network. This approach uses MLP layers as permutation-invariant functions to process each point of a point cloud individually, and a max-pooling layer to aggregate them. The performance of the network is limited because they do not consider local spatial relationships in the data. PointNet++ \cite{qi2017pointnet++} addresses this issue by sampling the points, grouping them in clusters, and applying PointNet on the clusters. SO-Net \cite{li2018so} uses a similar hierarchical structure adding self-organizing maps (SOMs) to capture better local structures. Other approaches like PointConv \cite{wu2019pointconv}, PointCNN \cite{li2018pointcnn} and KPConv \cite{thomas2019kpconv} construct kernels based on the input coordinates to be used as convolution weights. 

\noindent \review{\textbf{Projection-based approaches.}}
%separated this group from the previous paragraph
Some works project local neighborhoods into tangent planes and process them with 2D convolutions. The tangent plane parameters can be found using point tangent estimation \cite{tatarchenko2018tangent}, or approximated \cite{lin2020fpconv, huang2019texturenet, yang2020pfcnn}. The downside of these approaches is that they lose the information of 1 dimension given that they project the points to a local 2D plane. 
%Our work apply a similar hierarchical structure with a small variant, and use self-attention to improve the aggregation of local features.

\noindent \textbf{Self-attention and transformers.}
Self-attention and transformers have revolutionized the area of NLP \cite{vaswani2017attention, shaw2018self}. This has lead the 3D segmentation field to investigate these techniques \cite{guo2020pct, qiu2021semantic, zhao2020exploring}. In the point cloud domain, self-attention networks can be seen as an improvement of the MLP networks, where instead of capturing local relationships by using pooling layers \review{or weighting the features of the neighbors using hard-coded scores \cite{liu2020closer}}, they learn these relationships through an attention layer which estimates a score function to weight the contribution of each neighbor. The self-attention architecture resembles the encoder part of the transformers.
% Proposed to aggregate the features of each neighbor by the distance between the centroid point and its neighbors.
% \review{\cite{XXX} combines the features inside a patch of points by using the distance between the centroid and neighbor points as weights for the aggregation.} 
%\review{\cite{XXX} proposes a non-learnable attention mechanism that uses the distance between the centroid and neighbor points as weights (attention scores) for the local aggregation.}
One of the attempts to apply transformers to point clouds is PCT \cite{guo2020pct}. They replace the MLP layers from PointNet++ with transformer layers and the output feature of each layer is enriched with a discrete Laplacian operator. There are two differences between PCT and ours. The first one is that they aggregate the points inside a neighborhood by applying maxpooling. We use self-attention instead, which allows the information from all the neighbors to be passed, rather than only using the neighbor with the highest feature value. The second difference is that they use global attention and we use local attention. This means that their method attends to all points, while ours attends to a local cluster of points. In terms of computation, their memory requirements are quadratic $[N \times N]$ while ours is $[K \times N]$ where $N$ is the number of points and $K$ the number of neighbors in the local cluster. \cite{qiu2021semantic} uses geometric features to add extra information to semantic features, and self-attention and maxpooling are used to do the local aggregation. The same operation is performed at different scales and the results are combined using another self-attention layer. %\review{The way they process the features is the closest to ours. However, they combine the geometric and semantic features and obtain the neighborhood information using pooling operations, this makes them lose information about the local patch.} %do not uncomment(The way they perform local aggregation is the closest to ours. However, this design loses information from the geometric features by combining them indiscriminately with the semantic features.)
%Our network approaches this problem by using a two-head self-attention mechanism, where one head focuses on the geometric features and the other on the semantic features. \review{The self-attention mechanism also allows our network to learn how much each neighbor contributed to a given patch.} In addition, they use scalar attention while we use vector attention, the latter brings more flexibility to the attention layer. This is because scalar attention uses the same learned score for all the feature channels of a neighbor point, whereas vector attention obtains a score for each channel individually \cite{zhao2020exploring}. Our method can also be considered as the closest implementation to a transformer for point cloud segmentation, because, unlike previous works, we adapted one of the key properties of transformers, which is the multi-head attention structure.

\review{The goal of 3D point cloud segmentation is to find good local relationships. Most of the methods use pooling operations \cite{qi2017pointnet, qiu2021semantic} to extract the important features inside a patch. However, this loses some information about the neighbors because these operations either return the maximum or average value of a group of points. The neighborhood information can be preserved better by using an attention mechanism, which helps the network to learn how much each neighbor contributes to the local patch. Although there are approaches that use self-attention \cite{qi2017pointnet++, guo2020pct, qiu2021semantic}, the type of attention they use is \textit{scalar}. The problem with scalar attention is that it uses the same learned score for all the feature channels of a neighbor point. Our network instead uses vector attention \cite{zhao2020exploring}, which computes a score for each channel individually, bringing more flexibility to the layer. We extend this flexibility by introducing a two-headed self-attention layer, where one head focuses on geometric features and the other on semantic features. Even though there are works \cite{qiu2021semantic, lin2020fpconv} that use geometric and latent features, they treat them indistinguishably, losing the individual contribution of each feature. Our method can also be considered as one of the closest implementations to transformers for point cloud segmentation because, unlike previous research, we adapted one of the key properties of transformers, which is the multi-head attention structure.}

\section{Proposed Method}
\label{sec:prop_method}
\begin{figure}[t]
    \begin{center}
    %\fbox{\rule{0pt}{2in} \rule{0.9\linewidth}{0pt}}
       \includegraphics[scale=0.35]{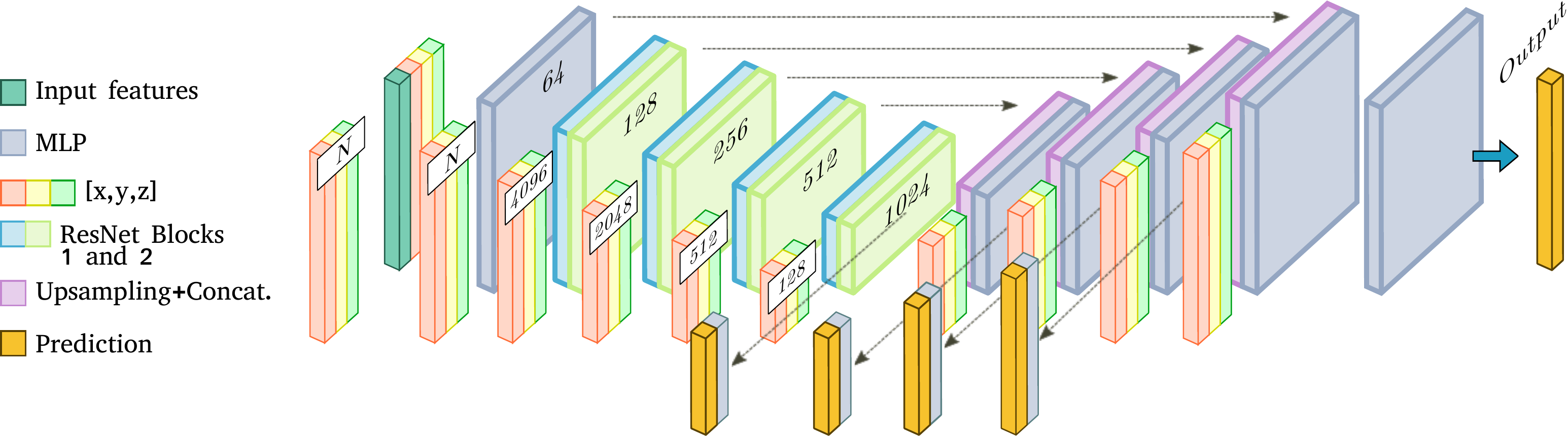}
    \end{center}
   \caption{The encoder-decoder architecture receives as input $xyz$ coordinates and $RGB$. The figure shows the effect of the sub-sample process on the $xyz$ values and the output features of each encoder layer. The encoder consists of ResNet blocks, which have our Ge-Latto layer (see Figure \ref{fig:resnet_blocks}). The decoder consists of up-sampling layers which are concatenated with their respective encoder features using residual connections and combined with an MLP.}
\label{fig:network}
\end{figure}

Ge-Latto extracts two types of information from the point cloud: The geometric information obtained from the Cartesian coordinates of the points and the latent feature information learned by the network each time the point cloud is sub-sampled. The first part of this subsection describes the network architecture and sampling strategy. The second part describes the two-headed attention layer and explains how latent and geometric information is used. 

\subsection{Network and Sampling}
\begin{figure}[t]
\begin{center}
%\fbox{\rule{0pt}{2in} \rule{0.9\linewidth}{0pt}}
   \includegraphics[scale=0.20]{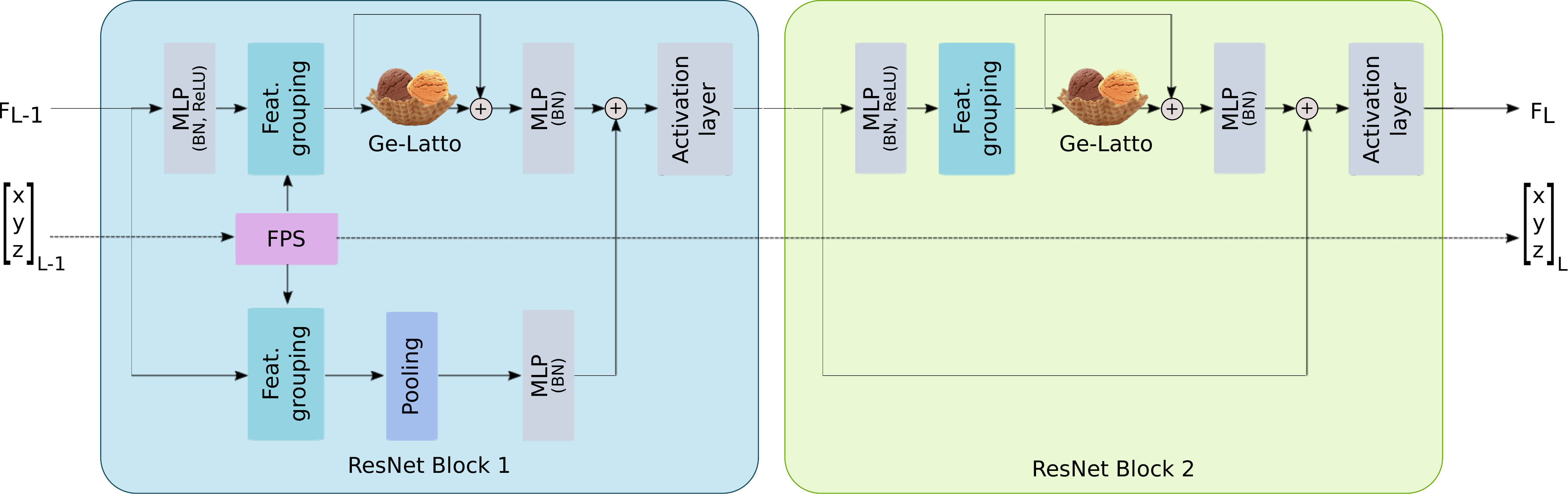}
\end{center}
   \caption{ResNet Blocks. The first block sub-samples the point cloud and finds nearest neighbors inside a radius between the sampled points and the input points. Because of the sampling, the residual connection has a maxpooling layer to match the input with the output size. The function of the second block is similar to the first one, but without sub-sampling.} %the points.} % removed
\label{fig:resnet_blocks}
\end{figure}

The network has an encoder-decoder architecture and receives as input the $xyz$ coordinates and RGB color. Those features are projected to a higher dimension using a shared MLP layer\footnote{In the paper, we use the word MLP to refer to a shared MLP layer with 1 hidden dimension.} (see Figure \ref{fig:network}). The encoder reduces the number of points and extracts high-level features from a neighborhood of points. For this, each layer sub-samples the number of points of its input. Therefore $N_{l} > N_{l+1}$ where $N$ is the number of points and $l$ is the layer. The encoder has 4 layers that are designed like bottleneck ResNet blocks \cite{he2016deep} with Ge-Latto replacing the 2D convolutions. Using Thomas \etal \cite{thomas2019kpconv} configuration, the input features of a ResNet block are processed by an MLP layer followed by batch normalization and ReLU. The other MLPs of the block are only followed by batch normalization (see Figure \ref{fig:resnet_blocks}). %(commented) for details). 
%The first block is in charge of sampling the input points and grouping the input points in neighborhoods using the sampled points as centers. Meanwhile, the second block search neighbors using only the sampled points; Figure \ref{fig:nbr_group} shows an example of this.

\begin{figure}[t]
    \begin{center}
    %\fbox{\rule{0pt}{2in} \rule{0.9\linewidth}{0pt}}
       \includegraphics[scale=0.4]{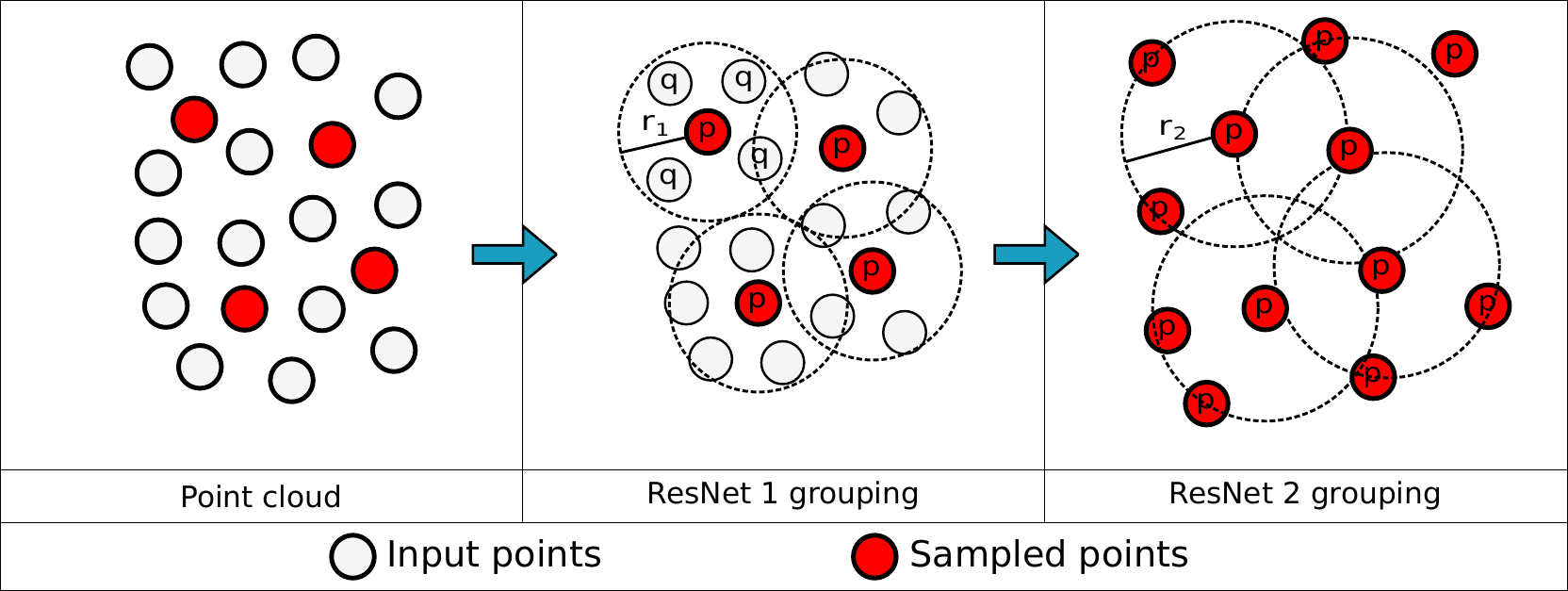}
    \end{center}
   \caption{Clustering process inside ResNet blocks. The first image shows the input points of the layer. The grouping criteria of Block 1 and 2 are shown in the second and third image. Block 1 groups the input points using the sampled points as centers with a radius $r_1$, whereas Block 2 does the grouping on the sampled points with a bigger radius $r_2$.}
\label{fig:nbr_group}
\end{figure}

Given the sparse nature of a point cloud, the choice of the sampling method is not trivial. The sampled points have to represent a group of points and be beneficial for the information ``aggregation'' of its neighbors. Here, we chose Farthest Point Sampling (FPS) because it outputs a more uniform-like distribution which is a desired property for point cloud semantic segmentation \cite{thomas2019kpconv}. 

The next step groups each point $p_i$ from the input set $\mathcal{P}_{l}$ (of size $N_l$) with their neighbors to find local-spatial relationships. The points can be either grouped by kNN or radius neighbors. We use the latter because it is more robust with non-uniform sampling settings like point clouds \cite{thomas2019kpconv}. Therefore, for each representative point $p_i\in \mathcal{P}_l$, $K$ points inside the given radius are randomly picked. We use $\mathcal{Q}_i$ to represent the grouped neighboring points of $p_i$ in the rest of the paper. It is important to note that $\mathcal{Q}_i \subseteq \mathcal{P}_{l-1}$, the only exception is in the second ResNet block, where $\mathcal{Q}_i \subseteq \mathcal{P}_{l}$ because no sampling is carried out; Figure \ref{fig:nbr_group} and Figure \ref{fig:resnet_blocks} show an example. The network also increases the receptive field by doubling the radius at every layer. 
%An example of this can be found in Figure \ref{fig:resnet_blocks}

For segmentation, the decoder up-samples the number of points until it recovers the size of the input of the network. The up-sampling of the features is done via tri-linear interpolation following Lin \etal \cite{lin2020fpconv}. The interpolated features are concatenated with the features from the corresponding encoder stage thanks to the residual connections (see Figure \ref{fig:network}). The final and auxiliary outputs of the decoder are feature vectors for each point in the input point set. An MLP is used to map these features to the final logits, whose feature dimension is the number of classes. The size of the auxiliary outputs corresponds to the number of points their respective layers have. The network has 4 auxiliary outputs, one for the last encoder layer, and three for the following decoder layers. For classification, global average pooling is used over the last encoder features to get a global feature vector of the point cloud. This feature is passed to an MLP to obtain the classification logits. 
%; the last layer is used for the final prediction.

\subsection{Two-headed Attention}
\begin{figure}[t]
\begin{center}
%\fbox{\rule{0pt}{2in} \rule{0.9\linewidth}{0pt}}
   \includegraphics[scale=0.35]{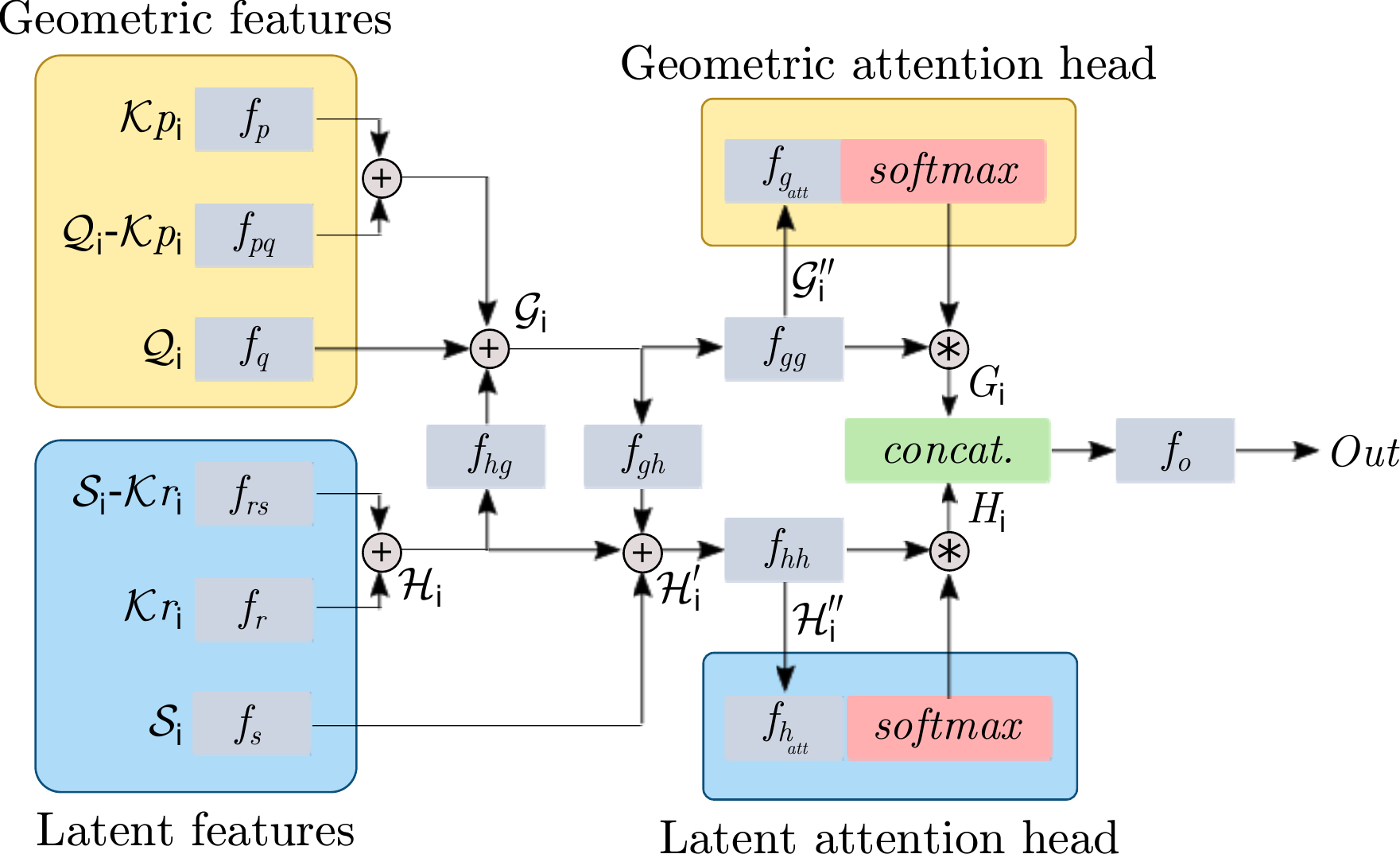}
\end{center}
   \caption{The two-headed Ge-Latto layer computes the local-attention for the geometric and latent features individually and then combines them using $f_i$ MLP layers. }
\label{fig:pla_layer}
\end{figure}

We claim that our two-headed attention layer finds better features by combining geometric and latent information in each layer of the encoder (Figure \ref{fig:pla_layer}). The geometric features that are used are the absolute position of the representative points $p_i \in  \mathbb{R}^{3}$, the $K$ neighbor points $\mathcal{Q}_i \in  \mathbb{R}^{K \times 3}$, and the relative position of the neighbors $\mathcal{Q}_i-\mathcal{K} p_i\in  \mathbb{R}^{K \times 3}$, where the operator $\mathcal{K}$ replicates the vector K times. 
%Here, each value of $\mathcal{Q}_i$ is subtracted by the center point $p_i$.
The latent information are the features learned by the hidden layers of the network. Each Ge-Latto layer uses those that belong to each centroid or representative point $r_i \in  \mathbb{R}^{D}$, the features of the neighbors $\mathcal{S}_i \in  \mathbb{R}^{K \times D}$, and the difference between the $K$ neighbors and centroids features $\mathcal{S}_i-\mathcal{K} r_i\in \mathbb{R}^{ K \times D}$, where $D$ is the dimensionality of the features, which is the number of feature planes shown in Figure \ref{fig:network}. The feature values are mapped linearly using an MLP layer $f_i$. The combined geometric and latent features are represented by $\mathcal{G}_i \in  \mathbb{R}^{K \times D}$ and $\mathcal{H}_i \in  \mathbb{R}^{K \times D}$ respectively, and are computed as follows (see Figure \ref{fig:pla_layer}): First, the latent features $r_i$ and $\mathcal{S}_i-\mathcal{K}r_i$ are transformed by MLPs and combined by vector addition: $\mathcal{H}_i=f_{r}(\mathcal{K}r_i)+f_{rs}(\mathcal{S}_i-\mathcal{K}r_i)$.
%\begin{equation} \label{eq:feat_cent_rel}
%H_f=f_{p_h}(p_{h})+f_{p_hq_h}(q_{h}-p_{h})\\
%\end{equation}

Then, the geometric features are combined. From Eq. \ref{eq:geo_cent_rel_nbr}, $f_{p}(\mathcal{K}p_i)$ and $f_{q}(\mathcal{Q}_i)$ encode the global geometric context in 3D space of the representative points and its neighbors. Meanwhile $f_{pq}(\mathcal{Q}_i-\mathcal{K}p_i)$ represents the local geometric context. We augment the geometric context by adding and projecting the latent feature $\mathcal{H}_i$. 
\begin{equation} \label{eq:geo_cent_rel_nbr}
\mathcal{G}_i=f_{p}(\mathcal{K}p_i)+f_{pq}(\mathcal{Q}_i-\mathcal{K} p_i)+f_{q}(\mathcal{Q}_i)+f_{hg}(\mathcal{H}_i)
\end{equation}

In the same way, the latent features are combined and augmented by adding the projected geometric feature $\mathcal{G}_i$. As Eq. \ref{eq:geo_cent_rel_nbr} encodes the geometric context, Eq. \ref{eq:feat_cent_rel_nbr_geo} encodes the latent context. $f_{r}(\mathcal{K}r_i)$ and  $f_{s}(\mathcal{S}_i)$ represent the global latent information  and $f_{rs}(\mathcal{S}_i-\mathcal{K}r_i)$ represents the local latent information.
\begin{equation} \label{eq:feat_cent_rel_nbr_geo}
\begin{split}
\mathcal{H}_i' & =\mathcal{H}_i+f_{s}(\mathcal{S}_i)+f_{gh}(\mathcal{G}_i)  =f_{r}(\mathcal{K}r_i)+f_{rs}(\mathcal{S}_i-\mathcal{K}r_i)+f_{s}(\mathcal{S}_i)+f_{gh}(\mathcal{G}_i)
%H_f' & =f_{p_h}(p_{h})+f_{p_hq_h}(q_{h}-p_{h})+f_{q_h}(q_{h})+f_{pf}(H_p)
\end{split}
\end{equation}
%As next step, the resultant features $H_p$ and $H_f'$ are projected by another MLP layer each one. Then, self-attention is used to combine the features inside the neighborhood patch $K$ (Eq. \ref{eq:geo_self_att} and Eq. \ref{eq:feat_self_att}). Here, vector attention is used instead of scalar attention. This allows the network to ``attend'' individual feature channels \cite{zhao2020exploring}. The attention part consist of an MLP layer followed by a Softmax function to obtain the weights on the features of the grouped points. Because vector attention is used is used, the dimension of the output of the attention layer is be ,$[N \times K  \times D]$ which is the same size as of the geometric and latent features $H_{p}''$ and $H_{f}''$

The resultant features $\mathcal{G}_i$ and $\mathcal{H}_i'$ are each projected by another MLP layer: $\mathcal{G}_i'' =f_{gg}(\mathcal{G}_i)$ and $\mathcal{H}_i'' =f_{hh}(\mathcal{H}_i')$. Then, self-attention is used to combine the features inside the neighborhood patch (Eq. \ref{eq:geo_self_att} and Eq. \ref{eq:feat_self_att}). The attention part consists of an MLP layer followed by a normalization function (Softmax) $\phi$ to obtain the weights of the neighbor features. Here, vector attention is used instead of scalar attention. This allows the network to ``attend'' to individual feature channels \cite{zhao2020exploring}. The dimension of the attention weights, the geometric features $\mathcal{G}_i''$ and latent features $\mathcal{H}_i''$ is $[K  \times D]$. Finally, to aggregate the local features, each neighbor feature $g_k \in \mathcal{G}_i''$ and  $h_k \in \mathcal{H}_i''$ is multiplied element-wise by its respective weight and then all the neighbors $k$ are summed; the outputs $G_i$ and $H_i$ have dimension $D$. In the supplementary material we show how our local aggregation is similar to a Graph NN \cite{battaglia2018relational}.
%the scores obtained by each head (geometric and latent) are multiplied element-wise (symbol circle-dot) by their respective features and summed along the $K$ (neighborhood) dimension. 
\begin{align}
\label{eq:geo_self_att} G_i& =\sum_{k =1}^{K}(\phi(f_{g_{att}}(g_k)) \odot g_k)\\
\label{eq:feat_self_att} H_i& =\sum_{k=1}^{K}(\phi(f_{h_{att}}(h_k)) \odot h_k)
\end{align}
%Where $g_k \in \mathcal{G}_i'$ and $h_k \in \mathcal{H}_i'$ are the $k$ neighbor features.
%\begin{align}
%\label{eq:geo_self_att} g_i& %=\sum_{k=1}^{K}(Softmax(f_{g_{att}}(\mathcal{G}_i'')) \odot %\mathcal{G}_i'')\\
%\label{eq:feat_self_att} h_i& %=\sum_{k=1}^{K}(Softmax(f_{h_{att}}(\mathcal{H}_i'')) \odot %\mathcal{H}_i'')
%\end{align}

%\begin{equation} \label{eq:geo_self_att}
%\begin{split}
%H_{p}''& =f_{gp}(H_{p}) \\
%\hat{H}_{p}& =\sum_{k=1}^{K}(Softmax(f_{p_{att}}(H_{p}'')) \odot H_{p}'') \\
%\hat{H}_{p}& =\sum_{K}^{}(\text{Score}\odot H_{p}'')
%\end{split}
%\end{equation}
%\begin{equation} \label{eq:feat_self_att}
%\begin{split}
%H_{f}''& =f_{gf}(H_{f}') \\
%\hat{H}_{f}& =\sum_{k=1}^{K}(Softmax(f_{f_{att}}(H_{f}'')) \odot H_{f}'') \\
%\hat{H}_{f}& =\sum_{K}^{}(\text{Score}\odot H_{f}')
%\end{split}
%\end{equation}
The output $O_i$ of the layer is obtained by concatenating and projecting the geometric and latent features: $O_i=f_{o}([G_i;H_i])$, where $[G_i;H_i] \in \mathbb{R}^{2D}$ and $f_o: \mathbb{R}^{2D} \mapsto \mathbb{R}^{D}$.
%\begin{equation} \label{eq:concat_output}
%H_{out}=f_{h}([\hat{H}_{p};\hat{H}_{f}])
%\end{equation}

In the transformers literature \cite{vaswani2017attention, shaw2018self}, the geometric features $f_p(\mathcal{K}p_i)$ and $f_{pq}(\mathcal{Q}_i-\mathcal{K}p_i)$, from Eq. \ref{eq:geo_cent_rel_nbr}, can be seen as absolute and relative positional encodings, respectively. Therefore, Eq. \ref{eq:geo_cent_rel_nbr} provides information about the absolute position of the points, and the relative position of the neighbor points with respect to the representative (centroid) points. Similarly, $f_{rs}(\mathcal{S}_i-\mathcal{K}r_i)$ from Eq. \ref{eq:feat_cent_rel_nbr_geo} can be seen as the \textit{key} and \textit{query} components in the transformers settings, where instead of using dot product as similarity function to obtain the relationship between two vectors, the values are subtracted. Each head in our attention mechanism can also be considered as a multi-head attention layer with number of heads $n=D$ or feature dimension $D'=1$ for each head; see supplementary material for demonstration. 

%This is demonstrated in the  supplementary material. %this can be demonstrate as follows.

\section{Experiments}
\label{sec:experiments}
Our method was evaluated using 3 datasets: ShapeNetPart \cite{wu20153d} for 3D object part segmentation,  Stanford Large-Scale 3D Indoor Spaces (S3DIS) \cite{armeni20163d} for 3D scene segmentation, and ModelNet40 \cite{yi2016scalable} for 3D shape classification. %The following hyperparameters were used to train our network.
%\subsection{Datasets}
%\noindent \textbf{ShapeNetPart \cite{wu20153d}:} is a collection of 16,681 3D point clouds with 16 categories, each one with 2 to 6 part labels. For a fair comparison, we used the standard train/test splits provided by the dataset.

%\noindent \textbf{Stanford Large-Scale 3D Indoor Spaces (S3DIS) \cite{armeni20163d}:} consists of six real large-scale indoor areas from three different buildings. Each area has rooms whose points are labeled with 13 classes (e.g. ceiling, floor, chair) and have color information. The number of points in one room varies between 0.5 million to 2.5 million, depending on its size. 

%\noindent \textbf{ModelNet40 \cite{yi2016scalable}:} The dataset consist of 12,311 3D meshes and their normal vectors classified into 40 categories. For the experiments, the points are processed following Thomas \etal \cite{thomas2019kpconv} for a fair comparison.
% \subsection{Implementation details}
\textbf{Implementation details:} The implementation was built using the public library PyTorch \cite{NEURIPS2019_9015}. Adam is used as optimizer with learning rate $1\mathrm{e}{-4}$, $\beta_1=0.9$, $\beta_2=0.98$, and $\epsilon = 1\mathrm{e}{-9}$. Cross-entropy with label smoothing is used as loss function for all the outputs. The final loss consists of the sum of 4 auxiliary losses and the main loss: \review{$\mathcal{L} = \alpha_1*\mathcal{L}_{aux_1} + \alpha_2*\mathcal{L}_{aux_2} + \alpha_3*\mathcal{L}_{aux_3} + \alpha_4*\mathcal{L}_{aux_4}+ \mathcal{L}_{main}$}. The influence of the auxiliary losses is weighted by \review{$\alpha_i$} because we are only interested in the final prediction, which is optimized by the main loss. Following %\cite{zhao2017pyramid} 
\review{the results of our ablation study, all the $\alpha_i=0.4$ in the experiments}. The encoder consists of one layer with size $N$, to process the input features, followed by 4 layers with sizes: 4096, 2048, 512, and 128; as seen in Figure \ref{fig:network}. The radius (receptive field) of the first encoder layer with ResNet blocks is $0.10m$ and it doubles at every layer. The number of neighbors is 32 for all the layers except for the last one which is 16. This is because the last layer has fewer points than the rest. All the MLP layers from the ResNet blocks (see Figure \ref{fig:resnet_blocks}) are followed by batch normalization and ReLU. The output layer before the prediction consists of an MLP layer with batch normalization and ReLU followed by a dropout with a probability of 0.5.
%The training was done end-to-end and the gradients of each loss are propagated across the whole network. 
All the experiments were done using a single RTX2080Ti with a batch size of 2. The data augmentation consists of scaling, flipping, rotating, and perturbing the points. For S3DIS, the color was augmented by switching the RGB channels and adding noise.

\subsection{Scene Segmentation}
\begin{figure}[t]
\begin{center}
%\fbox{\rule{0pt}{2in} \rule{0.9\linewidth}{0pt}}
   \includegraphics[scale=0.45]{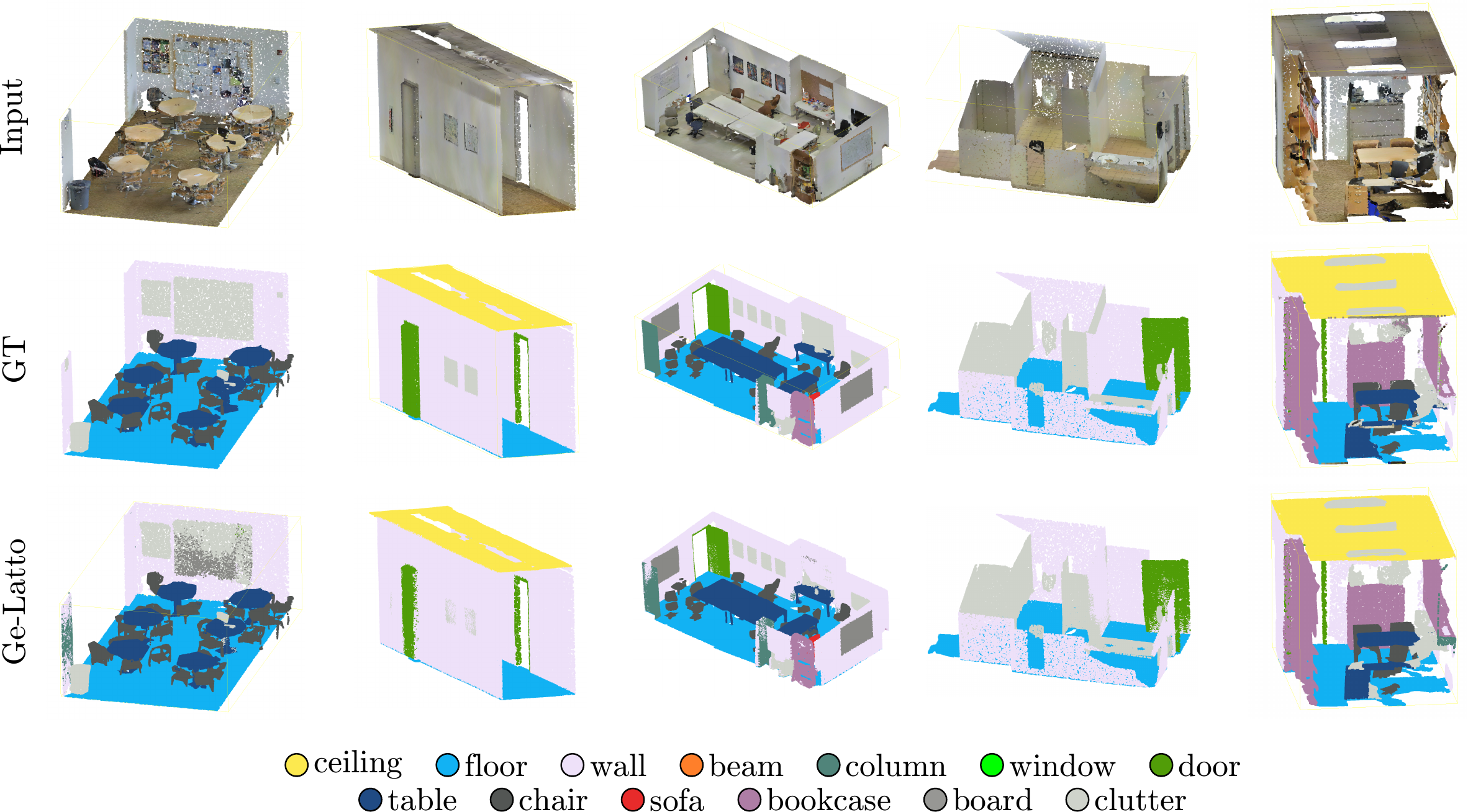}
\end{center}
   \caption{S3DIS results. }
\label{fig:s3dis_img}
\end{figure}

The S3DIS \cite{armeni20163d} dataset was used to test the network for scene segmentation. The dataset consists of six real large-scale indoor areas from three different buildings. Each area has rooms whose points are labeled with 13 classes (e.g. ceiling, floor, chair) and have color information. The number of points in one room varies between 0.5 million to 2.5 million, depending on its size. Because the number of points of each room is large, each room was split into blocks of size $[2m \times2m \times height]$. For training and testing, 6,144 points were randomly sampled and used as input. However, for testing, 6,144 points are randomly sampled until all the points inside a block are labeled. All the points are only sampled once, except when the total number of points inside a block is not a multiple of 6,144, in that case, the Softmax outputs are summed and the highest value is used as the predicted label. The evaluation metrics used are mean class-wise intersection over union (mIoU), mean of class-wise accuracy (mAcc), and overall accuracy (OA).
The dataset was evaluated in two ways: $1)$ Area 5 is used as test set and the network is trained using the other areas. $2)$ 6-fold cross-validation. Ge-Latto outperforms prior models in both evaluations. On area 5, it is $2.1$\% better than KPConv \cite{thomas2019kpconv} in mIoU (Table \ref{tab:comp_s3dis}), the qualitative results are shown in Figure \ref{fig:s3dis_img}. Meanwhile, on the k-fold cross-validation, it obtains the best OA (89.7\%), surpassing the previous state of the art of Qiu \etal \cite{qiu2021semantic} and obtaining better IoU in more objects (Table \ref{tab:comp_s3dis_kfold}).

\begin{table}
\centering
\small
 \setlength{\tabcolsep}{2.5pt}
 \resizebox{\textwidth}{!}{
\begin{tabular}{@{\extracolsep{4pt}}lccccccccccccccc} 
\hline
\multicolumn{1}{c}{Method}       & mIoU & mAcc & ceiling & floor & wall & beam & column & window & door & table & chair & sofa & bookcase & board & clutter  \\  
\cline{1-1} \cline{2-3} \cline{4-16}
PointNet \cite{qi2017pointnet}     & 41.1 & 49.0   & 88.8    & 97.3  & 69.8 & \textbf{0.1}  & 3.9    & 46.3   & 10.8 & 59.0    & 52.6  & 5.9  & 40.3     & 26.4  & 33.2     \\
SegCloud \cite{tchapmi2017segcloud}    & 48.9 & 57.4 & 90.1    & 96.1  & 69.9 & 0.0    & 18.4   & 38.4   & 23.1 & 70.4  & 75.9  & 40.9 & 58.4     & 13.0    & 41.6     \\
FPConv \cite{lin2020fpconv}      & 62.7 & 68.9 & \textbf{94.6}    & 98.5  & 80.9 & 0.0    & 19.1   & 60.1   & 48.9 & 80.6  & 88.0    & 53.2 & 68.4     & 68.2  & 54.9     \\
MinkowskiNet \cite{choy20194d} & 65.3 & 71.7 & 91.8    & 98.7  & \textbf{86.2} & 0.0    & \textbf{34.1}   & 48.9   & 62.4 & 81.6  & 89.8  & 47.2 & 74.9     & 74.4  & 58.6     \\
KPConv \cite{thomas2019kpconv}      & 67.1 & 72.8 & 92.8    & 97.3  & 82.4 & 0.0    & 23.9   & 58.0     & \textbf{69.0}   & 81.5  & 91.0    & 75.4 & \textbf{75.3}     & 66.7  & 58.9     \\
PCT \cite{guo2020pct} & 61.3 & 67.6 & 92.5   & 98.4  & 80.6 & 0.0    & 19.4  & 61.6     & 48.0   & 76.6  & 85.2    & 46.2 & 67.7     & 67.9  & 52.3    \\
Bilateral \cite{qiu2021semantic} & 65.4 & 73.1 & 92.9 & 97.9 & 82.3 & 0.0 & 23.1  & \textbf{65.5} & 64.9 & 78.5 & 87.5 & 61.4 & 70.7 & 68.7 & 57.2 \\
\hline
Ge-Latto (ours) &\textbf{69.2} & \textbf{75.9} & 94.5 & \textbf{99.2} & 84.0 & 0.0 & 24.5 & 56.3 & 68.9 & \textbf{84.2} & \textbf{92.4} & \textbf{82.8} & 70.9 & \textbf{76.9} & \textbf{64.6}          \\
\hline
\end{tabular}
}
\caption{S3DIS Area 5 results. The reported metrics are the mean class segmentation (mIoU), mean of class-wise accuracy (mAcc), and IoU for each class.}
\label{tab:comp_s3dis}
\end{table}
\begin{table}
\centering
\small
 \resizebox{\textwidth}{!}{
\begin{tabular}{@{\extracolsep{4pt}}lcccccccccccccccc} 
\hline
\multicolumn{1}{c}{Method} & OA            & mIoU          & mAcc          & ceiling       & floor         & wall          & beam          & column        & window        & door          & table         & chair         & sofa          & bookcase      & board         & clutter        \\ 
\cline{1-1} \cline{2-4} \cline{5-17}
PointNet  \cite{qi2017pointnet}                 & 78.5          & 47.6          & 66.2          & 88.0          & 88.7          & 69.3          & 42.4          & 23.1          & 47.5          & 51.6          & 54.1          & 42.0          & 9.6           & 38.2          & 29.4          & 35.2           \\
PointCNN \cite{li2018pointcnn}                   & 88.1          & 65.4          & \textbf{88.1} & 94.8          & \textbf{97.3} & 75.8          & 63.3          & 51.7          & 58.4          & 57.2          & 71.6          & 69.1          & 39.1          & 61.2          & 52.2          & 58.6           \\
SPGraph \cite{landrieu2018large}                   & -          & 62.1          & 73.0  & 89.9          & 95.1          & 76.4          & 62.8          & 47.1          & 55.3          & 68.4          & 73.5          & 69.2          & 63.2          & 45.9          & 8.7           & 52.9           \\
RandLA-Net \cite{hu2020randla}            & 88.0          & 70.0          & 82.0          & 93.1          & 96.1          & 80.6          & 62.4          & 48.0          & 64.4          & 69.4          & 69.4          & 76.4          & 60.0          & 64.2          & 65.9          & 60.1           \\
KPConv  \cite{thomas2019kpconv}                  & -             & 70.6          & 79.1          & 93.6          & 92.4          & \textbf{83.1} & 63.9          & \textbf{54.3} & \textbf{66.1} & \textbf{76.6} & 64.0          & 57.8          & \textbf{74.9} & \textbf{69.3} & 61.3          & 60.3           \\
Bilateral \cite{qiu2021semantic}                 & 88.9          & \textbf{72.2} & 83.1          & 93.3          & 96.8          & 81.6          & 61.9          & 49.5          & 65.4          & 73.3          & 72.0          & \textbf{83.7} & 67.5          & 64.3          & 67.0          & \textbf{62.4}  \\ 
\hline
Ge-Latto (ours)            & \textbf{89.7} & 71.4          & 81.3          & \textbf{95.3} & 95.1          & 82.3          & \textbf{69.2} & 51.9          & 64.8          & 73.3          & \textbf{77.3} & 59.6          & 71.1          & 63.0          & \textbf{67.4} & 57.9           \\
\hline
\end{tabular}
}
\caption{S3DIS dataset k-fold cross-validation comparison table.}
\label{tab:comp_s3dis_kfold}
\end{table}

\subsection{Object Part Segmentation}
\label{lbl:seg_part}
\begin{figure}[t]
\begin{center}
%\fbox{\rule{0pt}{2in} \rule{0.9\linewidth}{0pt}}
   \includegraphics[scale=0.6]{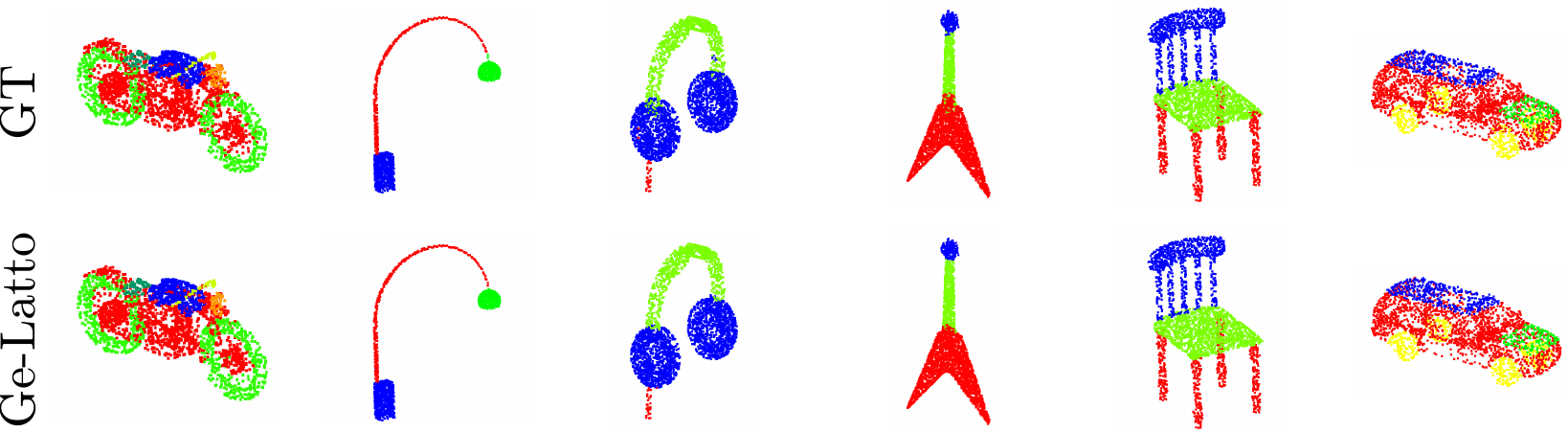}
\end{center}
   \caption{Our ShapeNetPart segmentation results. }
\label{fig:shapenet_img}
\end{figure}

The performance of our network in object part segmentation is measured by using the ShapeNetPart \cite{wu20153d} dataset. This dataset is a collection of 16,681 3D point clouds with 16 categories, each with 2 to 6 part labels. We used the standard train/test splits provided by the dataset. Category mean intersection over union (cat. mIoU) and instance mIoU are used as evaluation metrics. For training, 4096 points are randomly picked and used as input. For testing, the total number of points is used. The evaluation (Table \ref{tab:comp_shape_model}) shows that our model is only 0.9\% behind the current state of the art in cat. mIoU. Some of the wrong classifications are caused by noisy data, where some object components (e.g. rocket, motorbike, table) are wrongly labeled which is penalized by the metric. Figure \ref{fig:shapenet_img} shows some qualitative results.
\begin{table}
\small
\centering
\resizebox{.5\textwidth}{!}{
\begin{tabular}{@{\extracolsep{4pt}}lccc} 
\hline
\multicolumn{1}{c}{\multirow{2}{*}{Method}} & ModelNet40 & \multicolumn{2}{c}{ShapeNetPart}  \\ 
\cline{2-2}\cline{3-4}
\multicolumn{1}{c}{}                        & OA         & cat. mIoU & inst. mIoU               \\ 
\hline
PointNet \cite{qi2017pointnet}                                   & 89.2       & 80.4     & 83.7                   \\
PointNet++ \cite{qi2017pointnet++}                                  & 91.9       & 81.9     & 85.1                   \\
SO-Net  \cite{li2018so}                                    & 90.9       & 81.0     & 84.9                   \\
KPConv  \cite{thomas2019kpconv}                                     & 92.7       & \textbf{85.1}     & \textbf{86.4}                   \\
PCT  \cite{guo2020pct}                                     & \textbf{93.2}       & -    & \textbf{86.4}                   \\
\hline
Ge-Latto (ours)                             & 91.1       & 84.2     & 84.5                        \\
Ge-Latto (ours fine-tuned)                             &\textbf{93.2}       & -     & -                        \\
\hline
\end{tabular}
}
\caption{ModelNet40 and ShapeNet comparison table.}
\label{tab:comp_shape_model}
\end{table}

\subsection{Shape Classification}
The ModelNet40 \cite{yi2016scalable} dataset is used to study the performance of our network for shape classification. The dataset consists of 12,311 3D meshes and their normal vectors classified into 40 categories. For the experiments, the data is processed similar to Section \ref{lbl:seg_part}. For training, 7,168 points are randomly picked as input, and for testing all the points are used. The evaluation metric is overall accuracy. Our method achieves a competitive result of 91.1\% \review{when using the same hyper-parameters and network structure as ShapeNet. If the parameters are fine-tuned, the performance goes up to 93.2\%, matching the current best result. More details are given in the ablation study.} %, being only 1.6\% behind the best score.

\subsection{Ablation Study}
\label{sec:abliation}
%we moved all the ablation study from the supplementary material to the main paper
\noindent \textbf{Auxiliary losses.} The auxiliary losses provide a boost in performance to the network. Here, we consider that all weights $\alpha_i$ of the auxiliary losses have the same value and vary them from 0 to 1. Table \ref{tab:comp_abliation} shows that the best performance is obtained with $\alpha_i=0.4$, being $1.8\%$ (in terms of mIoU) better than the network trained without auxiliary losses. %Figure $XXX$ shows the prediction for each encoder layer.
\review{The supplementary material includes a more comprehensive study for different values of $\alpha_i$ on each loss term.} 

\begin{table}
\centering
\small
\resizebox{\textwidth}{!}{
\begin{tabular}{@{\extracolsep{4pt}}lcclcclccccc} 
\hline
\multicolumn{3}{c}{Auxiliary losses}               & \multicolumn{3}{c}{Attention heads}               & \multicolumn{3}{c}{Number of neighbors}      & \multicolumn{3}{c}{Features per multi-head} \\ 
%\hline
\cline{1-3} \cline{4-6} \cline{7-9}  \cline{10-12}
\multicolumn{1}{c}{Aux. weight $\alpha_i$} & mIoU & mAcc & \multicolumn{1}{c}{Attention heads} & mIoU & mAcc & \multicolumn{1}{c}{k} & mIoU & mAcc      &    \multicolumn{1}{c}{N Feat.} & mIoU & mAcc \\ 
\cline{1-3} \cline{4-6} \cline{7-9} \cline{10-12}
%\hline
$\alpha_i$ = 0.0                          & 67.4 & 73.2 & Only geometric                      & 66.3 & 72.0 & 8                     & 64.5 & 71.8       &                        1                     & \textbf{69.2} & \textbf{75.9}\\
$\alpha_i$ = 0.2                          & 68.5 & 74.4 & Only features                       & 66.5 & 72.3 & 16                    & 66.4 & 72.1       &                        2                     & 68.2 & 74.3\\
$\alpha_i$ = 0.4                          & \textbf{69.2} & \textbf{75.9} & Both                                & \textbf{69.2} & \textbf{75.9} & 32                    &\textbf{ 69.2} & \textbf{75.9}      &                        4                     & 68.0 & 74.2 \\
$\alpha_i$ = 0.6                          & 68.6 & 74.6 & MLP+pooling                         & 63.5 & 69.2 & 64                    & -    & -           &                        8                     & 68.0 & 74.1\\
$\alpha_i$ = 0.8                          & 68.3 & 74.3 & \multicolumn{1}{c}{}                &      &      & \multicolumn{1}{c}{}  &      &           \\
$\alpha_i$= 1.0                           & 68.1 & 74.0   & \multicolumn{1}{c}{}                &      &      & \multicolumn{1}{c}{}  &      &           \\
\hline
\end{tabular}
}
\caption{S3DIS ablation study experiments.}
\label{tab:comp_abliation}
\end{table}

\noindent \textbf{Two-headed attention.} Our proposed attention layer is evaluated using different variants: only with the geometric head, only with the feature head, both heads, and no heads (baseline). For the last one, an MLP+pooling layer replaces the attention mechanism. The experiments reported in Table \ref{tab:comp_abliation} show that only one head is enough to improve the baseline, and that the combination of both heads helps the network in the segmentation task.

\noindent \textbf{Number of neighbors.} Three networks with different neighborhood size were trained to find the number of neighbors that provide enough local information. The results from Table \ref{tab:comp_abliation} show that when $k \leq 16$, the number of neighbors might not be enough to provide a correct representation of the local context; $k=64$ could not fit in the GPU memory.

\noindent \textbf{Features per multi-head.} As shown in the supplementary material, each head (geometric and latent) of our attention layer is a special case of a multi-head attention with number of heads $n=D$ or feature dimension $D'=1$ per head. For this experiment, the number of features $D'$ per head was varied. More features per head means less heads (number of heads $n=D/D'$). In other words, $D'$ features will be weighted by the same attention score. The results from Table \ref{tab:comp_abliation} demonstrate that the more features a head has (less number of heads), the less flexible the network becomes. However, reducing the number of heads allows the network to be lighter, because it has to compute only $D/D'$ attention scores.

\noindent \textbf{Point cloud classification.} The model presented in Section \ref{sec:experiments} had the same network parameters for all the datasets to show that the proposed method can obtain competitive results without optimizing the hyper-parameters on each dataset. If the hyper-parameters are adjusted for a specific dataset, the performance of the network improves. In this section, the number of sub-sampling points per layer were modified following \cite{xiang2021walk}, where they mainly focus on point cloud classification. By sampling the points using the following values per layer, $N\rightarrow 1024 \rightarrow 512\rightarrow 256 \rightarrow 64$, instead of $N \rightarrow 4096  \rightarrow 2048\rightarrow 512\rightarrow 128$, the overall accuracy of our network increases from 91.1\% to 93.2\%, matching the state-of-the-art result. This seems to be caused by the number of sampled points and the global average pooling at the end of the encoder layer. Considering that the ModelNet40 dataset has objects %that share
\review{with} similar parts, like plants with flower pots, if one region of the object is bigger than others, this region will have more sampled points. Because the features of the sampled points at the last layer of the encoder are averaged, if there are more points representing a specific area, the averaged features will have a tendency to represent the wrong part of the object.

\noindent \textbf{Work flow video.} The video of our proposed method can be found at: \url{https://youtu.be/mjsttn3C89g}. \review{It shows how the point cloud is sampled during the encoder-decoder phase, the auxiliary outputs, and the learned geometric and latent attention weights.}

% \noindent \textbf{Work flow video.} The video\footnote{Following the double blind policy.} of our proposed method can be found at: \url{https://youtu.be/mjsttn3C89g}. \review{It shows how the point cloud is sampled during the encoder-decoder phase, the auxiliary outputs, and the learned geometric and latent attention weights.}

\section{Conclusion}
\label{sec:conclusion}
This paper proposes a novel two-headed attention mechanism capable of combining the geometric and latent information of neighbor points to learn richer features. This, combined with the leverage provided by the auxiliary losses, allow our network to work with real data and obtain the state of the art in the complex dataset S3DIS for 3D point cloud semantic segmentation. It also gets competitive results in the ShapeNetPart and ModelNet40 datasets.

\bibliography{egbib}

\end{document}

\maketitle

\begin{abstract}
This supplementary material is organized as follows: Section \ref{sec:multi-head} demonstrates that our geometric and latent heads are a special case of a multi-head attention layer in the transformers literature. Section \ref{sec:aux_outputs} shows an example of the prediction of each auxiliary output of the network. Section \ref{sec:gnn} describes the local aggregation step of our two-headed layer and how it relates to graph neural networks. In Section \ref{sec:attention_scores} the geometric and latent attention scores of our Ge-Latto layer are shown. Finally, Section \ref{sec:extended_abliation} extends the ablation study presented in the main paper. %(removed because it mas moved to the main paper) We also present a video that illustrates the semantic segmentation process of a point cloud. It shows, step by step how the point cloud is sampled during the encoder-decoder phase, the auxiliary outputs, and the learned geometric and latent attention weights of a point.
\end{abstract}
%-------------------------------------------------------------------------
\section{Special Case of Multi-head Attention Transformer}
\label{sec:multi-head}
%\iffalse
Each head (geometric or latent) in our attention mechanism can be considered as a multi-head attention layer with feature dimension (channels) $D'=1$ for each head or number of heads $n=D$, being $D$ the dimensionality of the features in each layer. This is demonstrated as follows.
\noindent Considering Eq. \ref{eq:attention}, the multi-head equation of a feature vector $H_i \in \mathbb{R}^ D$ proposed by Vaswani \etal \cite{vaswani2017attention}, where $Q_k$, $R_k$ and $V_k$ are the query, key and value vectors with size $D'$.

\begin{equation} \label{eq:attention}
\begin{split}
H_i&=\mathrm{Concat}(\mathrm{head_{i,1}}, \mathrm{head_{i,2}}, \mathrm{head_{i,3}},..., \mathrm{head_{i,n}}) \\
\mathrm{head_{i,n}}&=\sum_{k=1}^{K}((\mathcal{M} \phi(\gamma (Q_k, R_k)))\odot V_k) \\
\mathrm{where}: \\
\mathcal{M} &= \text{Replicates the vector $D'$ times} \\
\phi &= \text{Normalization function} \\
\gamma &= \text{Similarity function} \\
dim(\gamma(\cdot)) &= 1 \\
% dim(V_k) &= D' \\
dim(\mathrm{head_{i,n}}) &= D' \\
dim(H_i) &= D = n D' \\
K &= \text{Neighborhood size}
\end{split}
\end{equation}

Eq. \ref{eq:attention} shows there is one weight $\phi$ per head $n$, and each $\phi$ multiplies $D'$ feature channels. Therefore, \textbf{the feature $H_i \in \mathbb{R}^ D $ has $n$ weights $\phi$ and $nD'$ feature channels}. In the case where the number of heads $n$ is $D$, $D'$ would have the value of 1. There would be one weight $\phi$ per head, and each weight would multiply one feature channel. \textbf{This would give a vector $H_i$ with $D$ weights $\phi$ and $D$ feature channels}, which are the dimensions of our geometric and latent head features, as seen in Eq. \ref{eq:attention_d1} (latent feature equation), or Eq. 3 and Eq. 4 in the main paper.
%In Eq. \ref{eq:attention}, $Softmax(\cdot)$ outputs a score for each neighbor $k \in K$. This score is multiplied by each feature vector $D'$ of each point $k$ from $Value$. The size of $D'$ can vary between 1 and $D$, where $D$ is the size of the concatenated features $H$. In the case where $D'=D$, the number of heads is 1. Our attention layer is the special case when $D'=1$. Using $G_i$ as example we have:
%Each head has the size of $[N \times D]$ the all the features.
%In our case, $Value$ has a size of $[N \times K \times D]$, and the output of the $Softmax(\cdot)$ is $[N \times K \times 1]$. Therefore, the Softmax weights are
\begin{equation} \label{eq:attention_d1}
\begin{split}
H_{i}& =\sum_{k=1}^{K}(\phi(f_{h_{att}}(h_k)) \odot h_k) \\
%where: \\
%dim(f_{h_{att}}(h_k)) &= D \\ %d_n \\
%dim(h_k) &= D \\
%dim(H_i) &= D \\
%D' &= 1 \\
\end{split}
\end{equation}
Where the dimension of $f_{h_{att}}(h_k)$, $h_k$ and $H_i$ is $D$.
\section{Auxiliary Loss Outputs}
\label{sec:aux_outputs}
\begin{figure}[H]
\begin{center}
%\fbox{\rule{0pt}{2in} \rule{0.9\linewidth}{0pt}}
   \includegraphics[width=0.8\textwidth]{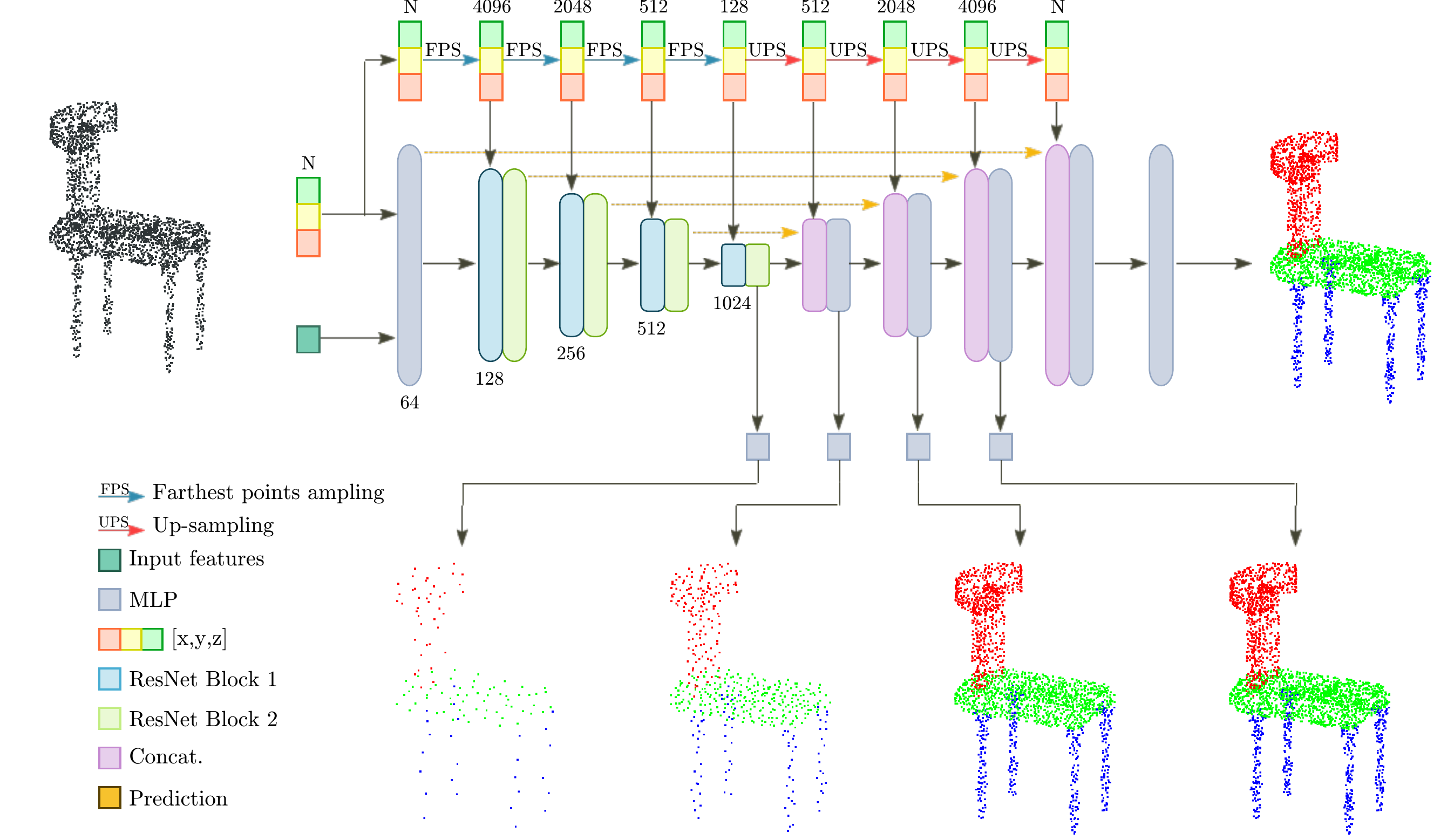}
\end{center}
   \caption{Auxiliary outputs. The network has 4 auxiliary outputs at multiple scales. One output comes from the last encoder layer and the other outputs plus the main output are obtained from the decoder layers.}
\label{fig:aux_loss}
\end{figure}

Each auxiliary loss optimizes the segmentation output of a sub-sampled version of the point cloud. Figure \ref{fig:aux_loss} shows an example of these outputs.

\section{Gelatto as a Graph Neural Network}
\label{sec:gnn}

Our network performs local aggregation following steps similar to those that a graph neural network (GNN) uses to update the value of a node through message passing: 
\begin{itemize}
    \item Our network updates the information of each node (center point $p_i$) based on the information from its neighborhood $q_k$, preserving graph symmetries (permutation invariance).
    \item The weights (edges) between the center point and its neighbors are learned through our geometric and latent self-attentions.
    \item At every step (each new layer of the network), the information from each node is propagated to a further (or neighbor) node until almost all the global information is aggregated. 
\end{itemize}

Table \ref{tab:gnn_gelatto} shows this relationship graphically.
\begin{table}[h!]
\centering
\begin{tabularx}{\textwidth}{cX}
\hline
\multicolumn{1}{c}{Illustration}  & \multicolumn{1}{c}{Local aggregation (message passing) steps}                                                                                                                                                                                                                                                                                                                                                                           \\ 
\hline
\raisebox{-1\totalheight}{\includegraphics[width=0.22\textwidth]{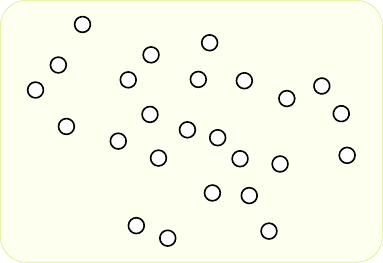}}

    & \begin{itemize}[leftmargin=*, topsep=0pt]
        \item Consider a set of points $p_i$ in the point cloud $P$.     
    \end{itemize}                                                                                                                                                                                                                                                                                                                                                                             \\
 
\raisebox{-1.\totalheight}{\includegraphics[width=0.22\textwidth]{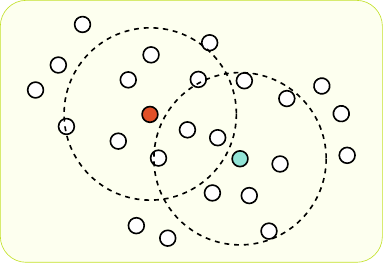}}

& \begin{itemize}[leftmargin=*, topsep=0pt]
    \item Our method obtains a local patch around a point by finding all the neighbor points inside a radius $r$. Here, we visualize the neighborhood of two points, red and cyan.                        
  \end{itemize} 
    \\
\raisebox{-1.\totalheight}{\includegraphics[width=0.22\textwidth]{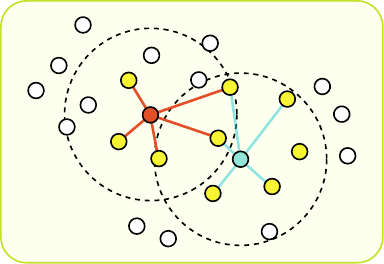}}

& \begin{itemize}[leftmargin=*, topsep=0pt]
    \item The patch is used to create a graph between the center points $p_i$ (red and cyan) and their neighbors $q_k$ (yellow). 
    \item To increase the robustness of our network, we randomly chose $k$ neighbors for each centroid. It can be seen as randomly zeroing the value of an edge.
\end{itemize}                                                                                                                                                                                                                                         \\
\raisebox{-1.25\totalheight}{\includegraphics[width=0.3\textwidth]{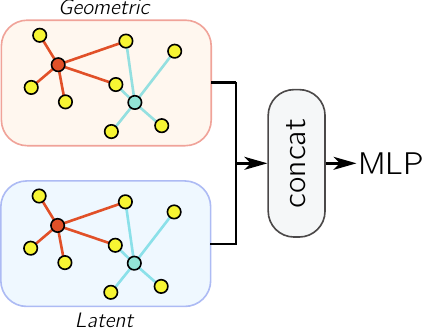}}
& 
\begin{itemize}[leftmargin=*, topsep=0pt]
    \item Our network uses self-attention (Eq. 3 and Eq. 4 in the main paper) to learn the edges that connect a centroid with its neighbors. 
    \item The information at a node is updated by aggregating the features from its neighbors and the learned edges. Gelatto creates two graphs, one for the geometric features, and another for the 
latent features. 
    \item After the \textit{message passing} step, the updated geometric and latent features are concatenated and processed by an MLP layer.
\end{itemize}
 
 \\
   
\hline
\end{tabularx}
% }
\caption{Local aggregation process of Gelatto.} %CHANGE COLOR of the first two figures
\label{tab:gnn_gelatto}
\end{table}

\section{Geometric and Latent Attention Scores}
\label{sec:attention_scores}
\begin{figure}[H]
\begin{center}
%\fbox{\rule{0pt}{2in} \rule{0.9\linewidth}{0pt}}
   \includegraphics[width=.8\textwidth]{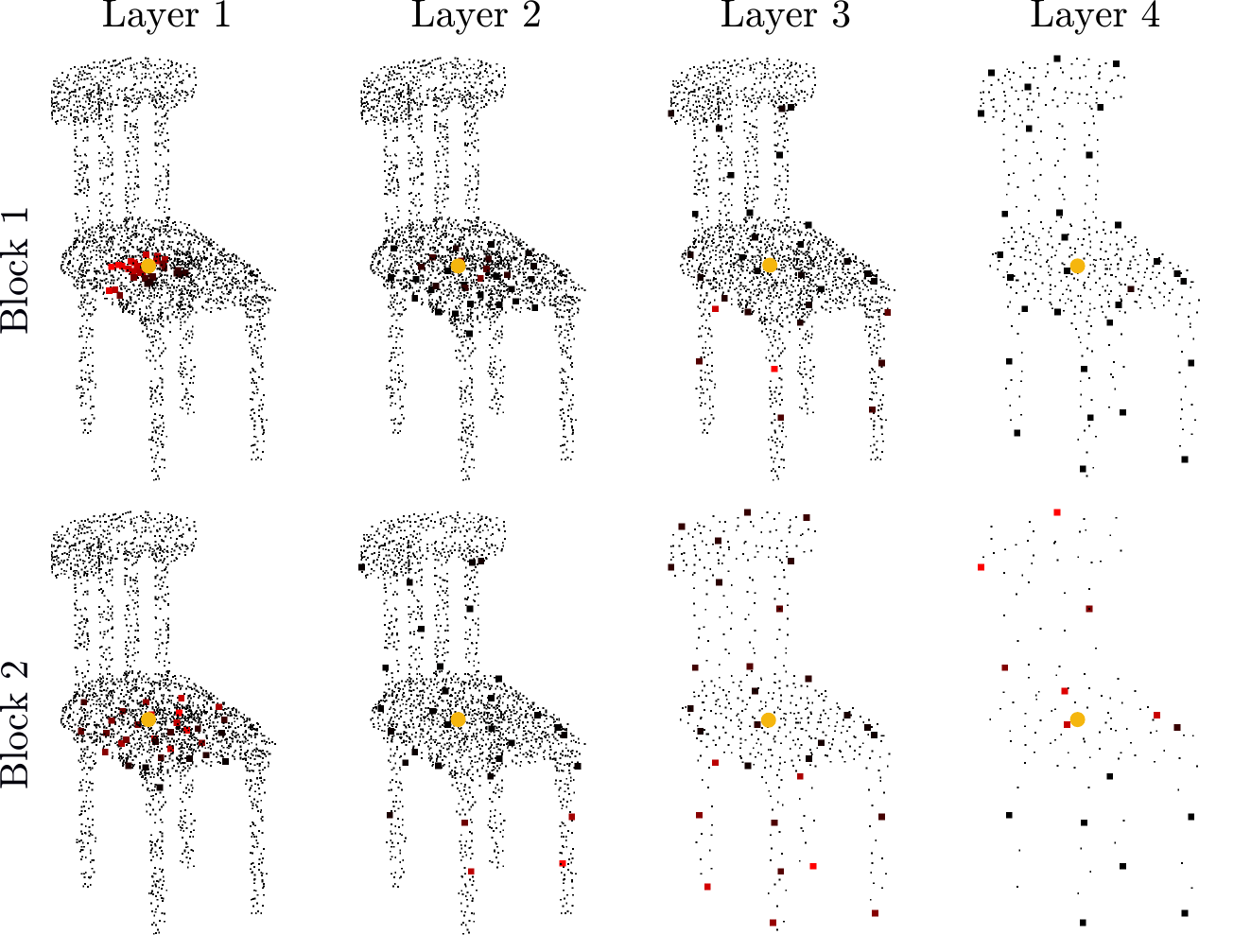}
\end{center}
   \caption{Learned geometric attention scores from a point (in yellow). The attention scores are represented in red, the stronger the intensity, the higher the score. The small black points are the sampled points at a given layer, the bigger points are the selected neighbor points inside a radius. The image shows the attention scores of the two ResNet Blocks at every encoder layer.}
   %      Each column shows the attention scores of an encoder layer and the rows shows the  
\label{fig:attention_scores_geo}
\end{figure}

\begin{figure}[H]
\begin{center}
%\fbox{\rule{0pt}{2in} \rule{0.9\linewidth}{0pt}}
   \includegraphics[width=0.8\textwidth]{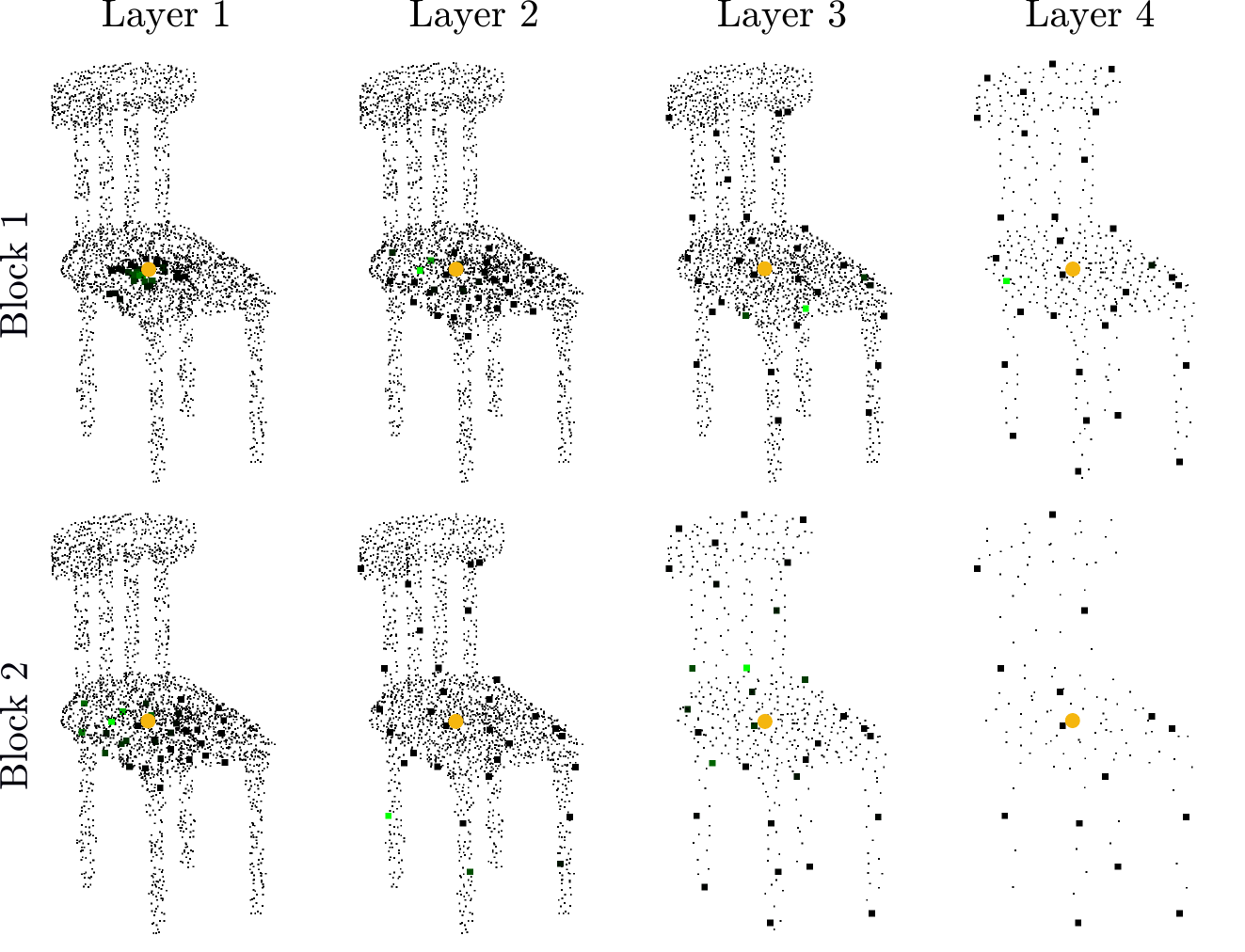}
\end{center}
   \caption{Learned latent attention scores from a point (in yellow). The attention scores are represented in green.}
\label{fig:attention_scores_lat}
\end{figure}

To show the attention scores learned by our Ge-Latto layer after each encoder step, an input point, that was not discarded by the sampling process, was picked. Because the attention score of each point has a dimension $D$, where $D$ is the dimensionality of a given layer, we randomly picked a value $d \in D$ per attention score to be shown in Figure \ref{fig:attention_scores_geo} and Figure \ref{fig:attention_scores_lat} for the geometric and latent heads, respectively. 

\begin{figure}[H]
\begin{center}
%\fbox{\rule{0pt}{2in} \rule{0.9\linewidth}{0pt}}
   \includegraphics[width=0.8\textwidth]{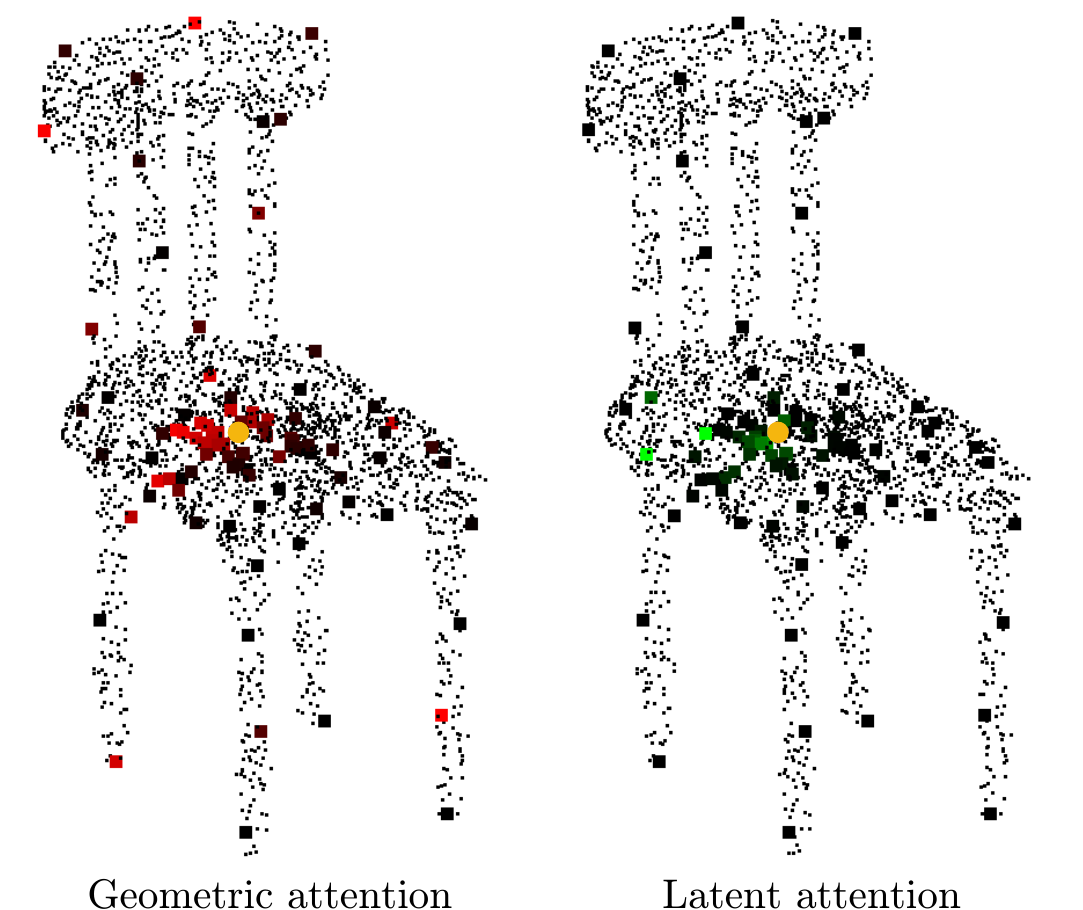}
\end{center}
   \caption{Geometric and Latent attention scores grouped.}
\label{fig:all_attentions}
\end{figure}

To observe how our network captures global and local relationships, all the attention scores of each head were grouped in Figure \ref{fig:all_attentions}. The figure shows that, for the chosen point, the latent head focuses more on closer points, meanwhile, the geometric head not only pays attention to the local points but also to more distant points, such as those of the back of the chair and legs. % like the back of the chair and legs. 

\section{Extended Ablation Study}
\label{sec:extended_abliation}
%A series of experiments are conducted to examine the important parts of our network: The auxiliary losses, two-headed attention and number of neighbors. The evaluation is done using the S3DIS dataset Area 5.
\subsection{Auxiliary losses}
\review{In the main paper, we only considered the case when all the auxiliary weights alpha $\alpha_i$ have the same value. This section explores different values for each alpha using grid search. The experiment consists of varying the alpha of one auxiliary loss, from 0 to 1 with increments of 0.2, and fixing the other alphas to 0.4 (the best value found before); the process is repeated for the 4 auxiliary losses. The values that the varying alpha can take are [0, 0.2, 0.6, 0.4, 0.8, 1], where 0 means that we do not minimize the loss for that output. We trained 24 variations of our network (4 auxiliary losses with 6 values for alpha). Each network was initialized with the weights from our best model, for the S3DIS area 5 dataset, and trained for 50 epochs. This experiment showed that there is no improvement when we vary the alphas individually and that the best value for this parameter is 0.4. However, we still observe that the use of auxiliary losses improves the performance of the network. As seen in Table 4 in the main paper, in the ablation study, when the network is trained without auxiliary losses, it obtains an IoU of $67.4\%$. Meanwhile, when the auxiliary losses are added, the performance increases to $69.2\%$.}

\subsection{Model Size and Speed}
\review{Table \ref{tab:param_s3dis} shows that our method achieves the state-of-the-art in the S3DIS dataset with fewer parameters than the previous methods. Our network has only 15.3M parameters, whereas KPConv has 25.8M, MinkowskiNet 21.7M, and FPConv 17.6M. The only network that has fewer parameters is PCT, 2.88M. However, we are $8\%$ better in point cloud semantic segmentation and obtain a similar performance in ModelNet40.} 
\begin{table}
\centering
\begin{tabular}{lcc} 
\hline
\multicolumn{1}{c}{Method} & Parameters     & mIoU           \\ 
\hline
MinkowskiNet               & 21.7M          & 65.3           \\
KPConv                     & 25.8M          & 67.1           \\
PCT                        & \textbf{2.88M} & 61.3           \\
FPConv                     & 17.6M          & 62.7           \\ 
\hline
Ge-Latto (ours)            & 15.3M          & \textbf{69.2}  \\
\hline
\end{tabular}
\caption{Model parameters comparison table. The parameters are in millions and the metric is for the S3DIS dataset.}
\label{tab:param_s3dis}
\end{table}

\review{The training and inference time using different numbers of points and batch sizes are shown in Table \ref{tab:speed_s3dis} reports. This table shows that it takes around 160ms to train 6144 points with a batch size of 2. At inference time, our network can analyze 6144 points with a batch size of 5 in 100ms, 20K points with a batch size of 3 in 210ms, and 200K points with a batch size of 3 in 300ms. All the tests were done using an NVIDIA RTX2080ti. These experiments show that our network is suitable for its applications that need a lighter network. }
\begin{table}
\centering
\begin{tabular}{lcccc} 
\hline
Number of points                     & 6K    & 10K   & 20K   & 200K   \\
\hline
Inference batch size                 & 5     & 3     & 3     & 3      \\
%\hline
Inference time                       & 100ms & 200ms & 210ms & 300ms  \\
%\hline
Training batch size                  & 2     & -     & -     & -      \\
%\hline
Training time                        & 160ms & -     & -     & -      \\
\hline
\end{tabular}
\caption{Training and inference time.}
\label{tab:speed_s3dis}
\end{table}

\bibliography{egbib}